\documentclass{llncs}
\usepackage{graphicx}
\usepackage{tabularx}
\usepackage{color}
\usepackage{url}
\usepackage{amsmath}
\usepackage{stmaryrd}
\usepackage{synttree}
\usepackage{tikz}
\usetikzlibrary{calc}

\newcommand{\acetext}[1]{\textsf{#1}}
\newcommand{\qacetext}[1]{``\textsf{#1}''}

\newcommand{\scopeopensymb}{\ensuremath{\mathord{\sslash}}}
\newcommand{\fwrefsymb}{\ensuremath{\mathord{>}}}

\newcommand{\bwrefsymb}{\ensuremath{\mathord{<}}}

\newcommand{\nrulesymb}[0]{\mathrel{:}}
\newcommand{\scrulesymb}[0]{\mathrel{\sim}}

\newcommand{\fs}[1]{\!\! \left( \! \scalebox{0.75}{$\begin{array}{l} \\[-2ex] #1 \\[-2ex] \end{array}$} \! \right)}
\newcommand{\nrule}[2]{#1 \: \xrightarrow{\displaystyle \: \nrulesymb \:} \: #2}
\newcommand{\scrule}[2]{#1 \: \xrightarrow{\displaystyle \: \scrulesymb \:} \: #2}

\newcommand{\scat}[1]{\:\: \mbox{\itshape #1} \:\:}
\newcommand{\cat}[2]{\:\: \mbox{\itshape #1} \, \fs{#2} }
\newcommand{\fwref}[1]{\:\: \fwrefsymb \fs{#1} }

\newcommand{\bwref}[1]{\:\: \bwrefsymb \fs{#1} }

\newcommand{\term}[1]{\:\: [\,\mbox{#1}\,] \:\:}

\newcommand{\preterm}[2]{\:\: \mbox{\underline{\itshape #1}} \, \fs{#2} }

\newcommand{\scopeopener}[0]{\:\: \scopeopensymb \:\:}

\newcommand{\featv}[2]{\mbox{#1:}\:\fboxsep 0.5mm \framebox{\scalebox{0.8}{#2}}\:\\}
\newcommand{\featc}[2]{\mbox{#1:}\:\mbox{#2}\\}

\newcommand{\tikzpoint}[1]{{\tikz[remember picture] \node (#1) {};}}

\begin{document}

\title{
  Codeco: A Grammar Notation for Controlled Natural Language in Predictive Editors
}

\author{
  Tobias Kuhn
}

\institute{
  Department of Informatics \& Institute of Computational Linguistics,\\
  University of Zurich, Switzerland\\
  \texttt{kuhntobias@gmail.com}\\
  \texttt{http://www.tkuhn.ch}
}

\maketitle

\begin{abstract}
Existing grammar frameworks do not work out particularly well for controlled natural languages (CNL), especially if they are to be used in predictive editors. I introduce in this paper a new grammar notation, called \emph{Codeco}, which is designed specifically for CNLs and predictive editors. Two different parsers have been implemented and a large subset of Attempto Controlled English (ACE) has been represented in Codeco. The results show that Codeco is practical, adequate and efficient.
\end{abstract}

\section{Introduction}

Controlled natural languages (CNL) like Attempto Controlled English (ACE) \cite{decoi2009rewerse} and others have been proposed to make knowledge representation systems more user-friendly \cite{schwitter2008owleddc,kuhn2009semwiki}. It can be very difficult, however, for users to write statements that comply with the restrictions of CNLs. Three approaches have been proposed so far to solve this problem: error messages \cite{clark2007kcap}, language generation \cite{power2009cnl}, and predictive editors \cite{schwitter2003eamtclaw,kuhn2009semwiki}. Each of the approaches has its own advantages and drawbacks. In my view, the predictive editor approach --- where the editor shows all possible continuations of a partial sentence --- is the most elegant one. However, implementing lookahead features (i.e. retrieving the possible continuations for a given partial sentence on the basis of a given grammar) is not a trivial task, especially if the grammar describes complex nonlocal structures like anaphoric references. Furthermore, good tool integration is crucial for CNLs, no matter which approach is followed. For this reason, it is desirable to have a simple grammar notation that is fully declarative and can easily be implemented in different kinds of programming languages.

These requirements are not satisfied by existing grammar frameworks. Natural language grammar frameworks like Head-driven Phrase Structure Grammars (HPSG) \cite{pollard1994hpsg} are unnecessarily complex for CNLs and do not allow for the implementation of efficient and reliable lookahead features. Grammar frameworks for formal languages --- called \emph{parser generators} --- like Yacc \cite{johnson1975yacc} do not allow for the declarative definition of complex (i.e. context-sensitive) structures. Definite clause grammars (DCG) \cite{pereira1986nlp} are a good candidate but they have the problem that they are hard to implement in programming languages that are not logic-based. Furthermore, all these grammar frameworks have no special support for the deterministic resolution of anaphoric references, as required by languages like ACE. The Grammatical Framework (GF) \cite{angelov2009cnl} is a recent grammar framework in competition with the approach to be presented here. It allows for declarative definition of grammar rules and can be used in predictive editors, but it lacks support for anaphoric references.

The following example illustrates the difficulty of lookahead in the case of complex nonlocal structures like anaphoric references. The partial ACE sentence \qacetext{every man protects a house from every enemy and does not destroy ...} can be continued by several anaphoric structures referring to earlier objects in the sentence: \qacetext{the man} or \qacetext{himself} can be used to refer to \qacetext{man}; \qacetext{the house} or \qacetext{it} can be used to refer to \qacetext{house}. It is not possible, however, to refer to \qacetext{enemy}, because it is under the scope of the quantifier \qacetext{every} and such references from outside the scope of a quantified entity do not make sense logically.\footnote{The difference between \qacetext{every man} and \qacetext{every enemy} in the given example is not obvious. The scope that is opened by the quantifier \qacetext{every} of \qacetext{every enemy} closes after \qacetext{enemy}, because the verb phrase ends there. In ACE, verb phrases --- but not only verb phrases --- close at their end all scopes that are opened within (which is, in general, also true for full natural English). Since \qacetext{every man} is not part of that verb phrase, the scope for \qacetext{every man} is still open at the end of the partial sentence. This is why \qacetext{man} is accessible for anaphoric references, but \qacetext{enemy} is not. See also Fig. \ref{fig:codecotree}.} Thus, a predictive editor should in this case propose \qacetext{the man} and \qacetext{the house} as possible continuations of the given partial sentence, but not \qacetext{the enemy}.

\section{Codeco}

In order to solve the problems sketched above, I created a new grammar notation called \emph{Codeco}, which stands for ``\emph{co}ncrete and \emph{de}clarative grammar notation for \emph{co}n\-trolled natural languages''. This grammar notation is designed specifically for CNLs to be used in predictive editors. Below is a brief description of the most important features of Codeco.

In a nutshell, grammar rules in Codeco consist of grammatical categories with flat feature structures\footnote{The restriction to flat feature structures (instead of general ones) restricts the expressive power of Codeco while making it easier to process and implement. The results suggest that Codeco is sufficiently expressive for common CNLs.}. One of the important features of Codeco is that anaphoric references and their potential antecedents can be represented by special categories for backward references ``\bwrefsymb'' and forward references ``\fwrefsymb'', respectively. These special categories establish nonlocal dependencies across the syntax tree in the sense that each backward reference needs a matching (i.e. unifiable) forward reference somewhere in the preceding text.

Furthermore, scopes can be defined that make the contained forward references inaccessible from the outside. In Codeco, scopes are opened by the special category ``\scopeopensymb'' and are closed by the use of scope-closing rules ``$\xrightarrow{\scrulesymb}$''. In contrast to normal rules ``$\xrightarrow{\nrulesymb}$'', scope-closing rules close at their end all scopes that have been opened by one of their child categories. In this way, Codeco allows us to define the positions where scopes open and close in a simple and fully declarative way. Other extensions that are not discussed here are needed to handle pronouns, variables, and references to proper names in the desired way. See \cite{kuhn2010phd} for the details.

The following grammar rules are examples that show how the special structures of Codeco are used:
\[\nrule{
  \scat{np}
}{
  \cat{det}{\featc{exist}{+}}\preterm{noun}{\featv{text}{Noun}}
  \fwref{\featc{type}{noun}\featv{noun}{Noun}}
}\]
\[\nrule{
  \scat{ref}
}{
  \term{the}\preterm{noun}{\featv{text}{Noun}}
  \bwref{\featc{type}{noun}\featv{noun}{Noun}}
}\]
\[\nrule{
  \cat{det}{\featc{exist}{--}}
}{
  \scopeopener
  \term{every}
}\]
\[\scrule{
  \cat{vp}{\featv{num}{Num}}
}{
  \cat{v}{\featv{num}{Num}\featc{type}{tr}}
  \cat{np}{\featc{case}{acc}}
  \scat{pp}
}\]
The first rule represents existentially quantified noun phrases like \qacetext{a house} and establishes a forward reference to define its status as a potential antecedent for subsequent anaphoric references. The second rule describes a definite noun phrase like \qacetext{the house} and declares its anaphoric property by a backward reference. The third rule describes the determiner \qacetext{every} opening a scope through the use of the scope opener category. The last rule describes a verb phrase and is scope-closing, which means that the end of such a verb phrase also marks the end of all scopes that have been opened within.

\begin{figure}[tp]
\begin{center}\scalebox{0.85}{
\branchheight{8mm}
\childsidesep{0.3em}
\childattachsep{0.3mm}
\synttree{8}
[s $\sim$
  [np
    [det
      [\tikzpoint{s1}]
      [.b \tikzpoint{l1}\acetext{Every}]
    ]
    [noun
      [.b \acetext{man}]
    ]
    [\tikzpoint{fw1}]
  ]
  [vp
    [vp \tikzpoint{t1}
      [v
        [tv
          [.b \acetext{protects}]
        ]
      ]
      [np
        [det
          [.b \acetext{a}]
        ]
        [noun
          [.b \acetext{house}]
        ]
        [\hspace{1mm} \tikzpoint{fw2}\hspace{1mm} ]
      ]
      [pp
        [prep
          [.b \acetext{from}]
        ]
        [np
          [det
            [\tikzpoint{s2}]
            [.b \tikzpoint{l2}\acetext{every}]
          ]
          [noun
            [.b \acetext{enemy}]
          ]
          [$>$]
        ]
      ]
    ]
    [conj
      [.b \tikzpoint{r1}\acetext{and}]
    ]
    [vp $\sim$
      [v
        [aux
          [\tikzpoint{s3}]
          [.b \tikzpoint{l3}\acetext{does not}]
        ]
        [tv
          [.b \acetext{destroy}]
        ]
      ]
      [np
        [ref
          [.b \acetext{...}]
          [\hspace{1mm} \tikzpoint{bw1}\hspace{1mm} ]
        ]
      ]
    ]
  ]
]
\begin{tikzpicture}[remember picture,overlay]
  \node at ($(l1) + (-0.2,0)$) (p1) {};
  \node at ($(l2) + (-0.2,0)$) (p2) {};
  \node at ($(l3) + (-0.2,0)$) (p3) {};
  \node at ($(r1) + (-0.15,0)$) (p4) {};
  \fill[fill opacity=0.1,blue] (t1 -| p2) rectangle (p4.south);
  \draw[very thick,blue!40] (s1) -| (p1);
  \draw[very thick,blue!40] (s2) -| (p2);
  \draw[very thick,blue!40] (s3) -| (p3);
  \draw[very thick,blue!40] (t1) -| (p4);
  \node[circle,inner sep=0.2mm,fill=blue!40] at (t1) (t1n) {$\sim$};
  \node[circle,inner sep=0.2mm,fill=blue!40] at (s1) (s1n) {$\scopeopensymb$};
  \node[circle,inner sep=0.2mm,fill=blue!40] at (s2) (s2n) {$\scopeopensymb$};
  \node[circle,inner sep=0.2mm,fill=blue!40] at (s3) (s3n) {$\scopeopensymb$};
  \node[circle,inner sep=0.5mm,fill=blue!40,draw=black] at (p1) (p1n) {\scriptsize $($};
  \node[circle,inner sep=0.5mm,fill=blue!40,draw=black] at (p2) (p2n) {\scriptsize $($};
  \node[circle,inner sep=0.5mm,fill=blue!40,draw=black] at (p3) (p3n) {\scriptsize $($};
  \node[circle,inner sep=0.5mm,fill=blue!40,draw=black] at (p4) (p4n) {\scriptsize $)$};
  
  \node[circle,inner sep=0.2mm,fill=red!40] at (fw1) (fw1n) {$>$};
  \node[circle,inner sep=0.2mm,fill=red!40] at (fw2) (fw2n) {$>$};
  \node[circle,inner sep=0.2mm,fill=red!40] at (bw1) (bw1n) {$<$};
  \draw[very thick,red!40,dashed] (fw1n) .. controls ($(fw1n) + (2,0)$) and ($(bw1n) + (-2,0)$) .. (bw1n);
  \draw[very thick,red!40,dashed] (fw2n) .. controls ($(fw2n) + (3,0)$) and ($(bw1n) + (-3,0)$) .. (bw1n);
\end{tikzpicture}
}\end{center}
\caption{This figure shows an exemplary syntax tree of a partial ACE sentence. Head nodes of scope-closing rules are marked with ``$\sim$''. The positions where scopes open and close are marked by parentheses. The shaded area marks a closed scope, which is not accessible from the outside. The dashed lines indicate the possible references.}
\label{fig:codecotree}
\end{figure}
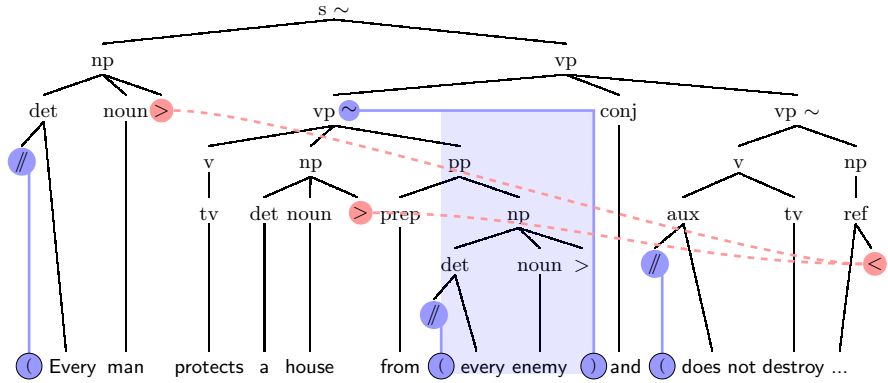

Codeco grammars allow for the correct and efficient implementation of lookahead features. Figure \ref{fig:codecotree} shows a possible syntax tree of the example above and visualizes how the possible anaphoric references can be found.

Two independent parsers have been implemented for Codeco: one that executes Codeco grammars as Prolog DCGs and another one that is a chart parser implemented in Java. These two implementations have been shown to be equivalent and efficient \cite{kuhn2010phd}. The chart parser for Codeco is used in the predictive editors of AceWiki \cite{kuhn2009semwiki} and the ACE Editor\footnote{\url{http://attempto.ifi.uzh.ch/webapps/aceeditor/}}. The DCG parser is used for regression and ambiguity tests for the grammars of both tools.

The Codeco grammar that is used in the ACE Editor consists of 164 grammar rules and describes a large subset of ACE covering nouns, proper names, intransitive and transitive verbs, adjectives, adverbs, prepositions, variables, plurals, negation, prepositional phrases, relative clauses, modality, numerical quantifiers, various kinds of coordination, conditional sentences, questions, anaphoric pronouns, and more. Currently unsupported ACE features are mass nouns, measurement nouns, ditransitive verbs, numbers and strings as noun phrases, sentences as verb phrase complements, Saxon genitive, possessive pronouns, noun phrase coordination, and commands. Nevertheless, this subset of ACE is --- to my knowledge --- the broadest subset of English that has ever been defined in a formal and fully declarative way and that includes complex issues like anaphoric references.

Codeco grammars can be checked for syntactic ambiguity by exhaustive language generation up to a certain sentence length. Due to combinatorial explosion, it is advisable not to check the complete language but to define a core subset thereof. Such a core subset should use a minimal lexicon and should have a reduced number of grammar rules, in a way that language structures that could possibly lead to complex cases of ambiguity are retained. For the ACE Codeco grammar introduced above, such a core subset has been defined, consisting of 97 of the altogether 164 grammar rules. Exhaustive language generation of this core subset up to ten tokens leads to 2'250'869 sentences, which are all different. Since syntactically ambiguous sentences would be generated more than once (because they have more than one syntax tree), this proves that at least the core of the ACE Codeco grammar is unambiguous, at least for sentences up to ten tokens. In the same way, it can be proven that the core of the ACE Codeco grammar is indeed a subset of ACE.

See my thesis \cite{kuhn2010phd} for the complete description of Codeco, the full ACE Codeco grammar, complexity considerations, the details and results of the performed tests and the involved algorithms.

\section{Conclusions}

In summary, the Codeco notation allows us to define controlled subsets of natural languages in an adequate way. The resulting grammars are fully declarative, can be efficiently interpreted in different kinds of programming languages, and allow for lookahead features, which is needed for predictive editors. Furthermore, it allows for different kinds of automatic tests, which is very important for the development of reliable practical applications.

Codeco is a proposal for a general CNL grammar notation. It cannot be excluded that extensions become necessary to express the syntax of other CNLs, but Codeco has been shown to work very well for a large subset of ACE, which is one of the most advanced CNLs to date.

\bibliography{codeco}

\begin{thebibliography}{10}

\bibitem{angelov2009cnl}
Krasimir Angelov and Aarne Ranta.
\newblock Implementing controlled languages in {GF}.
\newblock In {\em Proceedings of the Workshop on Controlled Natural Language
  ({CNL} 2009)}, volume 5972 of {\em Lecture Notes in Computer Science}, pages
  82--101. Springer, 2010.

\bibitem{clark2007kcap}
Peter Clark, Shaw-Yi Chaw, Ken Barker, Vinay Chaudhri, Phil Harrison, James
  Fan, Bonnie John, Bruce Porter, Aaron Spaulding, John Thompson, and Peter
  Yeh.
\newblock Capturing and answering questions posed to a knowledge-based system.
\newblock In {\em {K-CAP} '07: Proceedings of the 4th International Conference
  on Knowledge Capture}, pages 63--70. ACM, 2007.

\bibitem{decoi2009rewerse}
Juri~Luca De~Coi, Norbert~E. Fuchs, Kaarel Kaljurand, and Tobias Kuhn.
\newblock Controlled {E}nglish for reasoning on the {S}emantic {W}eb.
\newblock In {\em Semantic Techniques for the {W}eb --- The {REWERSE}
  Perspective}, volume 5500 of {\em Lecture Notes in Computer Science}, pages
  276--308. Springer, 2009.

\bibitem{johnson1975yacc}
Stephen~C. Johnson.
\newblock {Y}acc: Yet another compiler-compiler.
\newblock Computer Science Technical Report~32, Bell Laboratories, Murray Hill,
  NJ, USA, July 1975.

\bibitem{kuhn2009semwiki}
Tobias Kuhn.
\newblock How controlled {E}nglish can improve semantic wikis.
\newblock In {\em Proceedings of the Forth Semantic Wiki Workshop ({SemWiki}
  2009)}, volume 464 of {\em CEUR Workshop Proceedings}. CEUR-WS, 2009.

\bibitem{kuhn2010phd}
Tobias Kuhn.
\newblock {\em Controlled English for Knowledge Representation}.
\newblock Doctoral thesis, Faculty of Economics, Business Administration and
  Information Technology of the University of Zurich, Switzerland, to appear.

\bibitem{pereira1986nlp}
Fernando Pereira and David H.~D. Warren.
\newblock Definite clause grammars for language analysis.
\newblock In {\em Readings in Natural Language Processing}, pages 101--124.
  Morgan Kaufmann Publishers, 1986.

\bibitem{pollard1994hpsg}
Carl Pollard and Ivan Sag.
\newblock {\em Head-Driven Phrase Structure Grammar}.
\newblock Studies in Contemporary Linguistics. Chicago University Press, 1994.

\bibitem{power2009cnl}
Richard Power, Robert Stevens, Donia Scott, and Alan Rector.
\newblock Editing {OWL} through generated {CNL}.
\newblock In {\em Pre-Proceedings of the Workshop on Controlled Natural
  Language ({CNL} 2009)}, volume 448 of {\em CEUR Workshop Proceedings}.
  CEUR-WS, April 2009.

\bibitem{schwitter2008owleddc}
Rolf Schwitter, Kaarel Kaljurand, Anne Cregan, Catherine Dolbear, and Glen
  Hart.
\newblock A comparison of three controlled natural languages for {OWL} 1.1.
\newblock In {\em Proceedings of the Fourth {OWLED} Workshop on {OWL}:
  Experiences and Directions}, volume 496 of {\em CEUR Workshop Proceedings}.
  CEUR-WS, 2008.

\bibitem{schwitter2003eamtclaw}
Rolf Schwitter, Anna Ljungberg, and David Hood.
\newblock {ECOLE} --- a look-ahead editor for a controlled language.
\newblock In {\em Controlled Translation --- Proceedings of the Joint
  Conference combining the 8th International Workshop of the European
  Association for Machine Translation and the 4th Controlled Language
  Application Workshop ({EAMT-CLAW03})}, pages 141--150, Ireland, 2003. Dublin
  City University.

\end{thebibliography}
\bibliographystyle{plain}

\end{document}